\documentclass[
]{article}
\usepackage{a4wide}
\usepackage{palatino}
\usepackage{psfig}

\begin{document}

\title{Architectural Considerations for Conversational Systems --- \\
       The Verbmobil/INTARC Experience}
\author{G.\ G\"orz$^1$, J.\ Spilker$^1$, V.\ Strom$^2$, H.\ Weber$^1$ \\ 
        \mbox{}$^1$ University of Erlangen-Nuremberg, IMMD 
        (Computer Science) VIII --- AI \\
        Am Weichselgarten 9 \\
        D-91058 ERLANGEN \\
        Email: {\tt goerz@informatik.uni-erlangen.de} \\
        \mbox{}$^2$ University of Bonn, Institut of Communication Research 
        and Phonetics}
\date{}
\maketitle

\parskip 5pt
\parindent 0pt 

\section{Conversational Requirements for Verbmobil}

Verbmobil\footnote{This work was funded by the German Federal
Ministry for Research and Technology (BMFT) in the framework of the
Verbmobil Project under Grant BMFT 01 IV 101 H / 9. The
responsibility for the contents of this study lies with the authors.}
is a large German joint research project in the area spontaneous
speech-to-speech translation sytems which is sponsored by the German
Federal Ministery for Research and Education.  In its first phase
(1992--1996) ca.\ 30 research groups in universities, research
institutes and industry were involved, and it entered its second
phase in January 1997.  The overall goal is develop a system which
supports face-to-face negotiation dialogues about the scheduling of
meetings as its first domain, which will be enlarged to more general
scenarios during the second project phase.  For the dialogue
situation it is assumed that two speakers with different mother
tongues (German and Japanese) have some common knowledge of English.
Whenever a speaker's knowledge of English is not sufficient, the
Verbmobil system will serve him as a speech translation device to
which he can talk in his native language.

So, Verbmobil is a system providing {\it assistance\/} in conversations as
opposed to fully {\it automatic conversational\/} systems. Of course, it can
be used to translate complete dialogue turns.  Both types of
conversational systems share a lot of common goals, in particular
utterance understanding --- at least as much as is required to
produce a satisfacory translation ---, processing of spontaneous
speech phenomena, speech generation, and robustness in general.  A
difference can be seen in the fact that an autonomous conversational
system needs also a powerful problem solving component for the domain
of discourse, whereas for a translation system the amount of domain
knowledge is limited by the purpose of translation, where most of the
domain specific problem solving --- except tasks like calendrical
computations --- has to be done by the dialog partners.

A typical dialogue taken from the Verbmobil corpus is the following
one:

\begin{verbatim}
<SIL> GUTEN TAG HERR KLEIN 
<SIL> K-ONNEN WIR UNS AM MONTAG TREFFEN
<SIL> JA DER MONTAG PA-ST MIR NICHT SO GUT
<SIL> JA DANN TREFFEN WIR UNS DOCH AM DIENSTAG
<SIL> AM DIENSTAG HABE ICH LEIDER EINE VORLESUNG
<SIL> BESSER W-ARE ES BEI MIR AM MITTWOCH MITTAGS
<SIL> ALSO AM MITTWOCH UM ZEHN BIS VIERZEHN UHR HABE ICH ZEIT
<SIL> DANN LIEBER GLEICH NACH MEINEM DOKTORANDENTREFFEN
<SIL> WOLLEN WIR UNS NICHT LIEBER IN MEINEM B-URO TREFFEN
<SIL> NA JA DAS W-URDE GEHEN
<SIL> JA HERR KLEIN WOLLEN WIR NOCH EINEN TERMIN AUSMACHEN
<SIL> VIELLEICHT GINGE ES AM  MITTWOCH IN MEINEM B-URO
<SIL> DAS IST DER VIERZEHNTE MAI
<SIL> AM MITTWOCH DEN VIERZEHNTEN PA-ST ES MIR NICHT SO GUT
<SIL> AM DIENSTAG IN DIESER WOCHE H-ATTE ICH NOCH EINEN TERMIN
<SIL> ALSO DANN AM DIENSTAG DEN DREIZEHNTEN MAI
<SIL> VORMITTAGS ODER AM NACHMITTAG
<SIL> JA MACHEN SIE DOCH EINEN VORSCHLAG
<SIL> JA DANN LASSEN SIE UNS DOCH DEN VORMITTAG NEHMEN
<SIL> JA GUT TSCH-U-S
\end{verbatim}

{\tt <SIL>} = ``silence''.

\section{Prosody and Spontaneous Speech Phenomena}

To cope with spontaneous speech, prosody plays a decisive role.
Integration of prosody into a speech-to-speech translator as an
additional speech language interface is a current topic of research.
Within the Verbmobil project, the experimental system INTARC was
designed which performs simultaneous speech-to-speech translation
(cf.\ \cite{Goerz:1996,Amtrup:1997}).  In INTARC, particular emphasis
has been put on the issues of {\it incrementality\/} and (top-down)
{\it component interaction\/} in order to take into account
expectations and predictions from higher level linguistic components
for lower level components.  For this purpose time synchronous
versions of traditional processing steps such as word recognition,
parsing, semantic analysis and transfer had to be developed. In part
completely new algorithms had to be designed in order to achieve
sufficient processing performance to compensate for the lack of right
context in search. The use of prosodic phrase boundaries became
essential to reduce search space in parsing and semantic analysis.

A further goal was robustness: If a detailed linguistic analysis
fails, the system should be able to produce an approximately correct
output. For this purpose, besides the main data flow the system has a
second template-based transfer strategy as a supplement, where a rough
transfer is performed on the basis of prosodically focused words and a
dialogue act detection. 

Furthermore, various spontaneous speech phenomena like pauses,
interjections, and false starts are covered by INTARC's dialogue turn
based unification grammar (cf.\
\cite{Kasper_Krieger:KI-96,Kasper_Krieger:COLING-96}).

\section{Incremental, Interactive, and Time Synchronous Processing}

The general design goals of the INTARC system architecture were time
synchronous processing as well as incrementality and interactivity as
a means to achieve a higher degree of robustness and scalability.
Interactivity means that in addition to the bottom-up (in terms of
processing levels) data flow the ability to process top-down
restrictions considering the same signal segment for all processing
levels.  The construction of INTARC 2.0, which has been operational
since fall 1996, followed an engineering approach focussing on the
integration of symbolic (linguistic) and stochastic (recognition)
techniques which led to a generalization of the concept of a ``one
pass'' beam search.  Fig.\ 1, which is a screen shot of INTARC's user
interface, gives an overview of the overall system architecture.

\begin{figure}[ht]
    \begin{center}
      \leavevmode
    \end{center}
  \caption{The architecture of INTARC 2.0}
\end{figure}

To enable component interaction, we designed the communication
framework ICE \cite{Amtrup:1995,Amtrup:1996} which maps an abstract
channel model onto interprocess communication.  Its software basis is
PVM (Parallel Virtual Machine), supporting heterogeneous locally or
globally distributed applications.  The actual version of ICE runs on
four hardware platforms and five operating systems with interfaces to
eight programming languages or dialects.

\section{Interactions between Recognizer, SynParser, SemParser, and
Prosody}

To understand the operation of INTARC, we start with an overview of
its syntactic parser component (SynParser).  Whereas the dialogue
turn based grammar of the system is a full unification grammar
written in HPSG, SynParser uses only the (probabilistically trained)
context-free backbone of the unification grammar --- which
overgenerates --- {\it and\/} a context-sensitive probabilistic model
of the original grammar's derivations.  In particular, the following
preprocessing steps had to be executed:
\begin{enumerate}
  \item Parse a corpus with the original unification grammar $G$ to
        produce an ambiguous tree bank $B$.
  \item Build a stripped-down (type skeleton) grammar $G'$ such that
        for every rule $r'$ in $G'$ there is a corresponding rule $r$
        in $G$ and vice versa.
  \item Use an unsupervised reestimation procedure to train $G'$ on
        $B$ (context sensitive statistics).
\end{enumerate}

The syntactic parser (SynParser) is basically an incremental
probabilistic search engine based on \cite{Weber:1997} (for earlier
versions cf.\ \cite{Weber:1994a,Weber:1994b}); it receives word
hypotheses and phrase boundary hypotheses as input.  The input is
represented as a chart where frames correspond to chart vertices and
word hypotheses are edges which map to pairs of vertices. Word
boundary hypotheses (WBHs) are mapped to connected sequences of
vertices which lie inside the time interval in which the WBH has been
located. The search engine tries to build up trees according to a
probabilistic context free grammar supplied with higher order Markov
probabilities. Partial tree hypotheses are uniformly represented as
chart edges. The search for the $n$ best output trees consists of
successively combining pairs of edges to new edges guided by an
overall beam search strategy.  The overall score of a candidate edge
pair is a linear combination of three factors which we call decoder
factor, grammar factor and prosody factor. The decoder factor is the
well known product of the acoustic and bigram scores of the sequences
of word hypotheses covered by the two connected edges. The grammar
factor is the normalized grammar model probability of creating a
certain new analysis edge given the two input edges. The prosody
factor (see next section) is calculated from the acoustic WBH scores
and a class based tetragram which models sequences of words and phrase
boundaries.

So, SynParser performs purely probabilistic parsing without
unifications. Only $n$ best trees are transmitted to the semantic
parser component (SemParser) to be reconstructed deterministically
with unification.  SemParser uses a chart for representation and
reuse of partial analyses.  On failure, it issues a top-down request
to SynParser.  Because we make heavy use of structure sharing (to
depth $n$) for all chart edges we were able to achieve polynomial
runtime.  So, the main processing steps along the path recognizer ---
SynParser --- SemParser are the following:
  \begin{itemize}
    \item The {\it recognizer (decoder)\/} performs a beam search
          producing a huge lattice of word hypotheses.
    \item {\it SynParser\/} performs a beam search on this lattice to
          produce a small lattice of tree hypotheses.
    \item {\it SemParser\/} executes the unification steps in order
          to pick the best tree that unifies.
    \item Incremental bottom-up and top-down interaction of
          syntactic and semantic analysis are achieved by
          chart reconstruction and revision in SemParser.
    \item Furthermore, bottom-up input from recognizer is provided
          via a morphology module (MORPHY \cite{Althoff:1996}) for
          compound nouns.
  \end{itemize}
First experiments resulted in a runtime of approximately 30 times
real time (on a SuperSparc) and a recognition rate for {\it words in
valid trees\/} of approximately 50\%.  Current work is focussing
on fine tuning for word recognition, morphology, syntactic and
semantic parsing.

In the following we describe the interactions between the components
mentioned.
\begin{itemize}
  \item {\bf Interaction Recognizer--SynParser} (cf.\ \cite{Hauenstein:1994})
  \begin{itemize}
    \item The  (left-hand side connected) word graph is being transmitted
          by endpoints bottom up.
    \item Possible path extensions are being transmitted by starting
          points top down.
    \item This leads to the following {\it effects:\/}
          \begin{itemize}
            \item A dynamic modification of language perplexity for
                  recognition;
            \item Data reduction and search is being moved (partially)
                  from recognizer to parser.
          \end{itemize}
   \item Top-down interactions make only sense if there are strong
         model restrictions (narrow domain).
  \end{itemize}
  \item {\bf Interaction SynParser--SemParser} (cf.\ \cite{Kasper:KONVENS-96})
          \begin{itemize}
            \item Probabilistic Viterbi parsing of word graphs with $G'$ in
                  polynomial time (without unifications).
            \item Packing and transmission of $n$ best
                  trees(only trees with utterance status!)
                  per frame in $O($\#treenodes$)$ time complexity. \\
                  Protocol with powerful data compression.
            \item Trees are being reconstructed by SemParser by means of $G$
                  deterministically.  On failure a top-down request
                  for the next best tree is being issued.
            \item On failure, a top-down request for the next best tree is
                  being issued.
            \item Structure sharing (to depth $n$) for all edges
                  results in {\it polynomial runtime\/}.
            \item This yields a preference for the longest valid utterance.

          \end{itemize}
        A 100\% tree recognition rate results in unification grammar
        parsing in cubic time.

        So, in our case lattice parsing is tree recognition (decoding):
        \begin{itemize}
         \item For each new frame, a vertex and an empty agenda of search
               steps are created.
         \item All word hypotheses ending in the actual frame are read in
               as edges and all pairs of edges which can be processed are
               being scored and pushed on the agenda for that frame.
         \item The score is a weighted linear combination of log probability
               scores given by the models for acoustics, bigram, grammar and
               prosody.
         \item As in an acoustic beam recognizer all steps down to a given
               offset from the maximum score are taken and all others are
               discarded.
         \item The procedure stops when the word recognizer --- which
               supplies word hypotheses with acoustic scores --- sends an
               end of utterance signal.
        \end{itemize}
\end{itemize}

The interaction protocol implies that the first tree to be
transmitted is the best scored one: SynParser constructs its chart
incrementally, always sending the best hypotheses which have
utterance status.  SemParser reconstructs the trees incrementally and
reports failures.  While SemParser is working --- which may lead to a
rejection of this tree --- SynParser runs in parallel and finds some
new trees.  The failure messages are ignored as long as SemParser is
still constructing trees.  If SemParser becomes inactive, further
hypotheses with a lower score are sent.  SemParser utilizes its idle
time to reconstruct additional trees which may become important
during the analysis (``speculative evaluation'').  I.e., if the
estimation of an utterance improves over time, its subtrees are in
general not accessible to SemParser, since they have never got a high
score.  With speculative evaluation, however, we often find that they
have already been constructed, which helps to speed up parsing.
Since our grammar is turn-based, this situation is not the exception,
but in fact the normal case.  Hence, this strategy guarantees that
the utterance spanned by the trees increases monotonously in time.

A second phase is entered if SynParser has reached the end of the
word lattice.  In the case that SemParser has accepted one of the
previous trees as a valid reading, SynParser is being informed about
the success.  Otherwise SemParser calls for further tree hypotheses.
The selection criteria for the next best hypothesis are exactly the
same as in the first phase: ``Long'' hypotheses are preferred, and in
the case of equal length the one with the best internal score is
chosen.  I.e., in the second phase the length of a potential
utterance decreases.  If none of the requested trees are accepted, the
process stops iff SynParser makes no further trees available.  This
parameter controls the duration of the second phase.

Depending on the choice which trees are sent, SynParser directs the
behavior of SemParser.  This is the essential reason why SemParser
must not perform a search over the whole set of received hypotheses.
The stepwise reduction of the length of hypotheses guarantees that
the longest possible valid utterance will be found.  This is
particularly useful to analyze utterance parts when no fully spanning
reading can be found.

To summarize, the advantages of this protocol are that no search must
be performed by SemParser, that the best tree which covers the
longest valid utterance is being preferred (graceful degradation) and
that dynamic load-balancing is achieved.

\section{Issues in Processing Spontaneous Speech: Prosody and Speaker Style}

\subsection{Prosody}

The decisive role of prosody for processing spontaneous speech has
already been mentioned.  Now we describe the integration of prosodic
information into the analysis process from an architectural
viewpoint.  The {\bf interaction Parser--Prosody} can be summarized
as follows:
  \begin{itemize}
    \item Bottom-up hypotheses on the word boundary class are time
          intervals; they are attached incrementally to word lattice nodes.
    \item A prosodic score is computed from the word path, a trigram for words
          and phrase boundaries and an acoustic score for phrase
          boundaries (maximized).
    \item Prosody detectors are based on statistical classifiers,
          having been trained with prosodically labeled data.
    \item No use of word information is made;
          time assignment is done through syllable kernel detection.
    \item Recognition rates are: for accents 78\%, for phrase boundaries~81\%,
          and for sentence mood 85\%
  \end{itemize}

The prosody module consists of two independently working parts: the
phrase boundary detector \cite{Strom:1995} and the focus detector
\cite{Petzold:1995}.

The data material investigated consists of spontaneous spoken
dialogues on appointment scheduling. A subset of 80 minutes speech has
been prosodically labeled: Full prosodic phrases (B3 boundaries) are
distinguished from intermediate phrases (B2 boundaries). Irregular
phrase boundaries are labeled with B9, and the default label for a
word boundary is B0. The B2 and B3 boundaries correspond roughly to
the linguistic concept of phrase boundaries, but are not necessarily
identical (cf. \cite{Strom:1996}).

In the phrase boundary detector, first a parameterization of the
fundamental frequency and energy contour is obtained by calculating
eleven features per frame: F0 is interpolated in unvoiced segments
and decomposed by three band pass filters.  F0, its components,
and the time derivatives of those four functions yield eight F0
features which describe the F0 contour at that frame globally and
locally. Furthermore three bands of a short-time FFT followed by
median smoothing are used as energy features.

The phrase boundary detector then views a window of (if possible)
four syllables. Its output refers to the syllable boundary between
the second and the third syllable nucleus (in the case of a
4-syllable window). Syllables are found by a syllabic nucleus
detector based on energy features derived from the speech signal. For
each window a large feature vector is constructed. 

A Gaussian distribution classifier was trained to distinguish between
all combinations of boundary types and tones. The classifier output
was then mapped on the the four classes B0, B2, B3, and B9. The a
posteriori probabilities are used as confidence measure. When taking
the boundary with maximal probability the recognition rate for a test
set of 30 minutes is 80.76\%, average recognition rate is 58.85\%. 

The focus detection module of INTARC works with a rule-based
approach. The algorithm tries to solve focus recognition by global
description of the utterance contour, in a first approach represented
by the fundamental frequency F0.  A reference line is computed by
detecting significant minima and maxima in the F0 contour. The
average values between the maximum and minimum lines yield the global
reference line. Focus accents occur mainly in the areas of steepest
fall in the F0 course. Therefore, in the reference line the points
with the highest negative gradient were determined first in each
utterance. To determine the position of the focus the nearest maximum
in this region has been used as approximation. 

The recognition rate is 78.5\% and the average recognition rate is
66.6\%. The focus detection module sends focus hypotheses to the
semantic module and to the module for transfer and generation.
In a recent approach, phrase boundaries from the detector described
above where integrated in the algorithm.  After optimization
of the algorithm even higher rates are expected. 

As mentioned in the last section, one of the main benefits of prosody
in the INTARC system is the use of prosodic phrase boundaries inside
the word lattice search.

When calculating a prosody factor for an edge pair, we pick the WBH
associated with the connecting vertex of the edges. This WBH forms a
sequence of WBHs and word hypotheses if combined with the portions
already spanned by the pair of edges. Tests for the contribution of
the prosody factor to the overall search lead to the following
results: The same recognition performance in terms of $n$ best trees
could be achieved using 20\% less edges on the average. A lot of
edges are constant in a given search space --- namely those used for
the representation of the original set of word hypotheses and the
empty active rule edges which have a zero span. Counting only those
edges which are built up dynamically by the search process a
reduction of 65\% was measured.

In INTARC, the transfer module performs a dialog act based
translation. In a traditional deep analysis it gets its input (dialog
act and feature structure) from the semantic evaluation module. In an
additional path a flat transfer is performed with the best word chain
from the word recognition module and with focus information.

During shallow processing the focus accents are aligned to words. If a
focus is on a content word a probabilistically selected dialog act is
chosen. This dialog act is then expanded to a translation enriched
with possible information from the word chain.

Flat transfer is only used when deep analysis fails. First results
show that the `focus-driven' transfer produces correct --- but sometimes
reduced --- results for about 50\% of the data. For the other half of the
utterances information is not sufficient to get a translation; only 5\%
of the translations are absolutely wrong.. 

While the deep analysis uses prosody to reduce search space and
disambiguate in cases of multiple analyses, the `shallow focus based
translation' can be viewed as directly driven by prosody.

\subsection{Speaker Style}

A new issue in Verbmobil's second phase are investigations on
speaker style.  It is well known that system performance depends on
the perplexity of the language models involved.  Consequently, one of
the main problems is to reduce the perplexity of the models in
question.  The common way to approach this problem is to specialize
the models by additional knowledge about contexts. The traditional
n-gram model uses a collection of conditional distributions instead
of one single probability distribution. Normally, a fixed length
context of immediately preceding words is used. Since the length of
the word contexts is bound by data and computational resources,
practicable models could only be achieved by restricting the
application domain of a system. Commonly used $n$-gram models define
$P(w|C,D)$ where $C$ is a context of preceding words and $D$ is an
application domain. But also finer grained restrictions have been
tested in the last decade, e.g.\ a cache-based $n$-gram
\cite{Kuhn:1990}.

Intuitively, every speaker has its own individual speaking style. The
question is whether it is possible to take advantage of this fact.
The first step towards specialized speaker models is to prove whether
sets of utterances sorted by speakers show significant differences in
the use of syntactic structure at all.  So, first of all the whole
corpus has been tagged with POS-categories
grounded on syntactic properties of words (for tagger and
POS-categories see \cite{Schmid:1995}). Using the whole corpus, we
determined an empirical distribution $D_{all}$ over these categories.
In order to separate the corpus in typical and non typical speakers we
checked the distribution $D_s$ of every speaker $s$ against $D_{all}$
using the Chi-square test. While we can't say anything about the
usage of syntax by non-typical speakers, there is evidence that
typical speakers make a similar use of syntax in a rough sense. With
a significance level of 0.01 the test rejects 23.6\% of the speakers.

Bi- and trigram models were estimated on
the basis of the typical speakers and on the whole corpus in
comparison. On a test set of normal speakers only the specialized
models showed a slightly higher perplexity than the more general
models. In contrast to this the specialization explored with
automatic clustering using the K-means method shows a slightly better
perplexity on most of the test set speakers. As a distance measure
we take difference of two bigrams. The relatively small
improvement with specialized models is a result of the small amount of
data. Even partitioning of the corpus into few classes leads to
a lot of unseen pairs in the specialized bigrams. Hence a general
model trained on a larger amount of data could produce better results.

Using the results of the experiments above as a guideline we chose a
clustering procedure using a different clustering criterion. The
procedure is adapted from automatic word clustering
\cite{ueberla:1994,martin:1995}. The goal of the procedure is to find
a partitioning such that the perplexity of the specialized models is
minimized. To reduce the parameter problem we used a class-based $n$-gram
instead of the word-based bigram. Class-based $n$-grams estimates the
probability of a word sequence $w_1 \dots w_n$ by 
\[ \prod_{i=1}^n P(w_i|C(w_i))*P(C(w_i)|C(w_{i-1})) \mbox{  \hspace*{0.5cm}bigram
  class model}\]
or
\[ \prod_{i=1}^n P(w_i|C(w_i))*P(C(w_i)|C(w_{i-2})C(w_{i-1})) \mbox{\hspace*{0.5cm}
   trigram class model}\]
where $C(w)$ denotes the class of word $w$. $P(w_i|C(w_i))$ is called the
lexical part and {$P(C(w_i)|C(w_{i-1}))$} resp. $P(C(w_i)|C(w_{i-2})C(w_{i-1}))$ the grammatical part of the
model. We performed three different experiments to get an expression how
a speaking style affects the lexical and grammatical part:
\begin{enumerate}
\item 2POS test: $P(w_i|C(w_i))$ is assumed to be invariant. Only the grammatical part
  $P(C(w_i)|C(w_{i-1}))$ is adapted to every cluster. 
\item 3POS test: $P(w_i|C(w_i))$ is assumed to be invariant. Only the grammatical part
  $P(C(w_i)|C(w_{i-1})C(w_{i-2}))$ is adapted to every cluster. 
\item POS/word: Both parts are considered. 
\end{enumerate}
First clustering tests showed good results:
\begin{table}[htb]
\centering
\begin{tabular}{|l|r|} \hline
& Reduction\\ \hline
2POS& 6.5\% \\  \hline
3POS& 1.9\% \\ \hline
POS/Word & 10\% \\ \hline
\end{tabular}
\caption{Reduction of test set perplexity}
\end{table}
The best result was achieved by adapting both parts of the class
model. This fact corresponds with the intuitive expectation that
speaking style influences the selection of words and grammar rules.



\section{Recognition Results for INTARC 2.0}

For INTARC 2.0, a series of experiments has been carried out in order
to also compare empirically an incremental and interactive system
architecture with more traditional ones and to get hints for tuning
individual components and their interactions.

Basically, we tested three different module configurations:
\begin{description}
\item[DM] Decoder, Morphy (acoustic word recognition)
\item[DMP] Decoder, Morphy, Lattice Parser (word recognition in parsed
  utterances)
\item[DMPS] Decoder, Morphy, Lattice Parser, Semantic Module (word
  recognition in understood utterances)  
\end{description}
These configurations correspond to successively harder tasks, namely to
recognize, to analyze and to ``understand''.

We used the NIST scoring program for word accuracy to gain comparable
results. By doing this we gave preference to a well known and
practical measure although we know that it is in some way inadequate.
In a system like INTARC 2.0, the analysis tree is of much higher
importance than the recovered string. With the general goal of
spontaneous speech translation a good semantic representation for a
string with word errors is more important than a good string with a
completely wrong reading.  Because there does not yet exist a tree bank
with correct readings for our grammar, we had no opportunity to
measure something like a ``tree recognition rate'' or ``rule
accuracy''.

The word accuracy results in DMP and DMPS can not be compared to word
accuracy as usually applied to an acoustic decoder in isolation, whereas
the DM values can be compared in this way.  In DMP and DMPS we counted
only those words as recognized which could be built into a valid parse
from the beginning of the utterance. Words to the right, which could
not be integrated into a parse, were counted as deletions --- although
they might have been correct in standard word accuracy terms. Our
evaluation method is much harder than standard word accuracy, but it
appears to be a good approximation to ``rule accuracy''.  What cannot
be parsed is being counted as an error.  The difference between DMP
and DMPS is that a tree produced by the statistical approximation
grammar can be ruled out when being rebuilt by unification operations
in semantic processing. The loss in recognition performance from DMP
and DMPS corresponds to the quality of the statistical approximation.
If the approximation grammar had a 100\% tree recognition, there would
be no gap between DMP and DMPS.

The recognition rates of the three configurations were measured in
three different contexts. The first row shows the rates of normal
bottom-up processing. In the second row, the results of the phrase
boundary detector are used to disambiguate for syntax and semantics.
The third row shows the results of the system in top-down mode; here
no semantic evaluation is done because top-down predictions only
affect the interface between SynParser and Recognizer.

\begin{center}
\begin{tabular}{|l|c|c|c|} \hline 
         & DM  & DMP &  DMPS \\ \hline 
Word Accuracy & 93.9\% & 83.3\% &  47.5\% \\ \hline 
WA with phrase boundary & 93.9\% & 84.0\% & 48.6\% \\ \hline
WA in TD-Mode & 94.0\% & 83.4\% & --\\ \hline
\end{tabular}
\end{center}

\subsection{Conclusions}

Splitting composite nouns to reduce the recognizer lexicon shows good
results.  Search and rebuilding performed by the morphology module is
implemented as a finite state automaton, so there is no great loss in
performance.  Incremental recognition is as good as as the standard
decoding algorithms, but the lattices are up to ten times larger.  This
causes a performance problem for the parser. So we use an
approximation of an HPSG-Grammar for search such that syntactic
analysis becomes more or less a second decoding step. By regarding a
wider context, we even reduce the recognition gap between syntax and
semantics in comparison with our previous unification-based syntax
parser (see \cite{Weber:1994a,Weber:1994b}).  For practical usability
the tree-recognition rate must be improved. This can be achieved with a
bigger training set.  The dialogues we used contained only 83
utterances. Further improvement can be achieved by a larger context
during training to get a better approximation of the trees built by
the unification grammar.

Prediction of words seems to have no influence on the recognition
rate.  This is a consequence of the underlying domain.  Since the HSPG
grammar is written for spontaneous speech, nearly every utterance
should be accepted.  The grammar gives no restrictions on possible
completions of an utterance.  Restrictions can be only obtained by a
narrow beam-bound when compiling the prediction table.  But this leads
to a lower recognition rate because some correct words are pruned.

{\bf Acknowledgements.} We are grateful to all our colleagues within
the Verbmobil subproject on ``Architecture'' from the universities of
Bielefeld, Bonn, Hamburg, and from DFKI Saarbr\"ucken without whose
contributions within the last four years this article could not have
been written.


\begin{thebibliography}{99}

\bibitem{Althoff:1996}
Althoff, F., Drexel, G., L\"ungen, H., Pampel, M., and Schillo, Ch.:
{\it The Treatment of Compounds in a Morphological
Component for Speech Recognition.\/}  In: Gibbon, D.\ (Ed.):
{\it Natural Language Processing and Speech Technology. Results of
the 3rd KONVENS Conference\/}, Berlin: Mouton de Gruyter, October 1996

\bibitem{Amtrup:1995} 
Amtrup, J.: {\it ICE ---INTARC Communication Environment:
User's Guide and Reference Manual. Version 1.4.\/}  Verbmobil
Technical Document~14, Univ. of Hamburg, December 1995

\bibitem{Amtrup:1996} 
Amtrup, J., Benra, J.: {\it Communication in large distributed
AI systems for natural language processing.\/}  Proc.\ of COLING-96,
Kopenhagen, August 1996, 35--40

\bibitem{Amtrup:1997} 
Amtrup, J., Drexel, G., G\"orz, G., Pampel, M., Spilker, J.\ and Weber, H.:
{\it The parallel time-synchronous speech-to-speech system INTARC 2.0.\/}
Submitted to ACL-97

\bibitem{Carter:1994}
Carter, D.: {\it Improving Language Models by Clustering Training
Sentences.\/} Proc.\ of ANLP '94, Stuttgart, Germany, 1994. Extended version
in http://xxx.lanl.gov/cmp-lg/

\bibitem{Goerz:1996} 
G\"orz, G., Kesseler, M., Spilker, J. and Weber, H.:
{\it Research on Architectures for Integrated Speech/Language Systems in
Verbmobil.\/} Proc.\ of COLING-96, Kopenhagen, August 1996

\bibitem{Hauenstein:1994}
Hauenstein, A., Weber, H.: {\it An investigation of tightly
coupled time synchronous speech language interfaces.\/}
Proceedings of KONVENS-94, Vienna, Austria. Berlin: Springer,
September 1994

\bibitem{Kasper_Krieger:KI-96} 
Kasper, W. and Krieger, H.-U.: {\it Integration of prosodic and
grammatical information in the analysis of dialogs\/}. In:
G\"orz, G., H\"olldobler, S. (Ed.): {\it Proceedings of the 20th
German Annual Conference on Artificial Intelligence, KI-96, Dresden}.
Berlin: Springer (LNCS) 1996

\bibitem{Kasper_Krieger:COLING-96} 
Kasper, W. and Krieger, H.-U.: {\it Modularizing codescriptive
grammars for efficient parsing.\/} Proc.\ of COLING-96, Kopenhagen,
August 1996, 628--633.

\bibitem{Kasper:KONVENS-96}
Kasper, W., Krieger, H.-U., Spilker J., and Weber, H.: {\it From
word hypotheses to logical form: An efficient interleaved approach.\/}
In: Gibbon, D.\ (Ed.):
{\it Natural Language Processing and Speech Technology. Results of
the 3rd KONVENS Conference\/}, Berlin: Mouton de Gruyter, October 1996,
77--88.

\bibitem{Kuhn:1990} 
Kuhn, R. and DeMori, R: {\it A cache-based natural language model for
speech recognition.\/} IEEE Transactions on Pattern Analysis and Machine
Intelligence, 12(6), June 1990, 570--583 

\bibitem{martin:1995}
Martin, S., Liermann, J., and Ney, H.:{\it Algorithms for bigram and
  trigram word clustering. \/} Eurospeech '95, Madrid, Spain, 1995, 1253--1256.

\bibitem{Petzold:1995}
Petzold, A.: {\it Strategies for focal accent detection in
spontaneous speech.\/} Proc.\ 13th ICPhS Stockholm, Vol.\ 3, 1995, 672--675 

\bibitem{Schmid:1995}
Schmid, H.: {\it Improvements in Part-of-Speech Tagging with an Application
to German.\/} http://www.ims.uni-stuttgart.de/Tools/DecisionTreeTagger.html,
1995.

\bibitem{Strom:1995} 
Strom, V.: {\it Detection of accents, phrase boundaries and sentence
modality in German with prosodic features.\/} Proc.\ EUROSPEECH-95,
Madrid, 1995, 2039--2041 

\bibitem{Strom:1996}
Strom, V.: {\it What's in the `pure' prosody?\/} Proc.\ ICSLP 96,
Philadelphia, 1996

\bibitem{ueberla:1994}
Ueberla, J.P.: {\it An Extended Clustering Algorithm for Statistical Language Models},
  1994, Nr.: 9412003 E-Print Archive: http://xxx.lanl.gov/cmp-lg/.

\bibitem{Weber:1994a}
Weber, H.: {\it Time Synchronous Chart Parsing of Speech Integrating
Unification Grammars with Statistics.}
Speech and Language Engineering, Proceedings of the Eighth
Twente Workshop on Language Technology, (L.\ Boves, A.\ Nijholt, Ed.),
Twente, 1994, 107--119

\bibitem{Weber:1994b}
Weber, H.: {\it LR-inkrementelles probabili\-sti\-sches Chartparsing
von Worthypothesenmengen mit Unifikationsgrammatiken: Eine enge
Kopplung von Suche und Analyse.} Ph.D. Thesis, University of Hamburg, 1995,
Verbmobil Report 52.

\bibitem{Weber:1997}
Weber, H., Spilker, J., G\"orz, G.\ (1997): {\it Parsing N Best Trees
from a Word Lattice.} In: Nebel, B.\ (Ed). {\it Advances in
Artificial Intelligence. Proceedings of the 21st German Annual
Conference on Artificial Intelligence, KI-97, Freiburg}. Berlin:
Springer (LNCS) 1997

\end{thebibliography}
\end{document}